\documentclass[letterpaper, 10 pt, conference]{ieeeconf}  

\IEEEoverridecommandlockouts                              

\usepackage{graphics} 
\usepackage{epsfig} 
\usepackage{mathptmx} 
\usepackage{times} 
\usepackage{amsmath} 
\usepackage{amssymb}  
\usepackage{color}
\usepackage{multirow}
\usepackage{hhline}
\usepackage{cite}
\usepackage{caption}
\usepackage{url}
\usepackage{color,soul}
\usepackage{tablefootnote}
\usepackage[caption = false]{subfig}

\makeatletter
\let\NAT@parse\undefined
\makeatother
\usepackage{hyperref}

\title{\LARGE \bf
Stereo Visual Inertial LiDAR Simultaneous Localization and Mapping
}

\author{Weizhao Shao, Srinivasan Vijayarangan$^*$, Cong Li$^*$, and George Kantor
\thanks{$^*$These authors contributed equally to the paper}
\thanks{The authors are with the Robotics Institute, Carnegie Mellon University, Pittsburgh, PA 15213, USA. \{weizhaos, svijaya1, congl1, gkantor\}@andrew.cmu.edu}
}

\begin{document}
\maketitle
\thispagestyle{empty}
\pagestyle{empty}

\begin{abstract}
Simultaneous Localization and Mapping (SLAM) is a fundamental task to mobile and aerial robotics. LiDAR based systems have proven to be superior compared to vision based systems due to its accuracy and robustness. In spite of its superiority, pure LiDAR based systems fail in certain degenerate cases like traveling through a tunnel. We propose Stereo Visual Inertial LiDAR (VIL) SLAM that performs better on these degenerate cases and has comparable performance on all other cases. VIL-SLAM accomplishes this by incorporating tightly-coupled stereo visual inertial odometry (VIO) with LiDAR mapping and LiDAR enhanced visual loop closure. The system generates loop-closure corrected 6-DOF LiDAR poses in real-time and 1cm  voxel dense maps near real-time. VIL-SLAM demonstrates improved accuracy and robustness compared to state-of-the-art LiDAR methods.
\end{abstract}
\section{Introduction}

SLAM solves the problem of mapping unknown environments while estimating robot state. Though SLAM is actively researched for the past few decades, Cadena et al. \cite{pastPresent} note that there are still challenges in handling diverse environments and long-term continuous operations. SLAM systems operate on a wide range of sensor modalities each trying to exploit their benefits. In the past few years, LiDAR based SLAM systems have gained popularity over vision based systems due to their robustness to changes in the environment. However pure LiDAR based systems have their deficiencies. They fail in environments with repeating structures like tunnels or hallways. These environments are challenging to map and localize, and system which exploits the strengths of all the sensor modalities need to be deployed to succeed. We propose VIL-SLAM, which uses IMU, stereo cameras and LiDAR, and exploit their benefits collectively. Our experiments demonstrate that VIL-SLAM performs on par with pure LiDAR based systems in most cases and better on cases where pure LiDAR based systems simply fail. VIL-SLAM achieves this by integrating stereo VIO and LiDAR mapping with loop closure. To the best of our knowledge, this is the first work of this kind. In addition, we introduce a method to evaluate mapping results using a time-of-flight laser scanner (Faro). We also provide VIO validation results on the EuRoC MAV dataset.

VIL-SLAM uses a tightly-coupled stereo VIO that performs fixed-lag pose graph optimization, LiDAR mapping that uses sparse 3D features for map registration, and loop closure that integrates sparse point cloud alignment with visual loop detection. Loop closure optimizes a global pose graph using an incremental solver. VIL-SLAM is designed to operate long term and in different environments robustly. The high frequency IMU measurements produce estimates which are reasonable for the short interval but quickly drift. When constrained with stereo visual measurements, we can correct the biases and estimate accurate relative motion (referred to as VIO). The relative motion estimate is used to aid LiDAR scan matching which then accumulates the high-fidelity 3D point clouds to form an accurate map. The robot's state estimate accumulates drift during long traversals. Loop closure addresses this issue by recognizing the revisited sites using either visual or LiDAR methods. Visual methods involve using Bag-of-Words \cite{refBoW} to recognize the place and Perspective-n-Point (PnP) algorithm to estimate the pose correction. In LiDAR methods, the places are recognized using segment based algorithms like SegMatch \cite{segMatch}, and pose correction is estimated using Iterative Closest Point (ICP) \cite{121791} algorithm. While the Bag-of-Words method is fast and versatile, it lacks the accuracy of the slow but robust LiDAR method which uses ICP. VIL-SLAM uses a hybrid approach where it first finds the loop closure candidate using Bag-of-Words technique, generates a rough estimate of the pose correction using Perspective-n-Point (PnP) algorithm, and then refines the rough estimate using ICP. 

\begin{figure}[t!]
    \centering
    \includegraphics[scale=0.145]{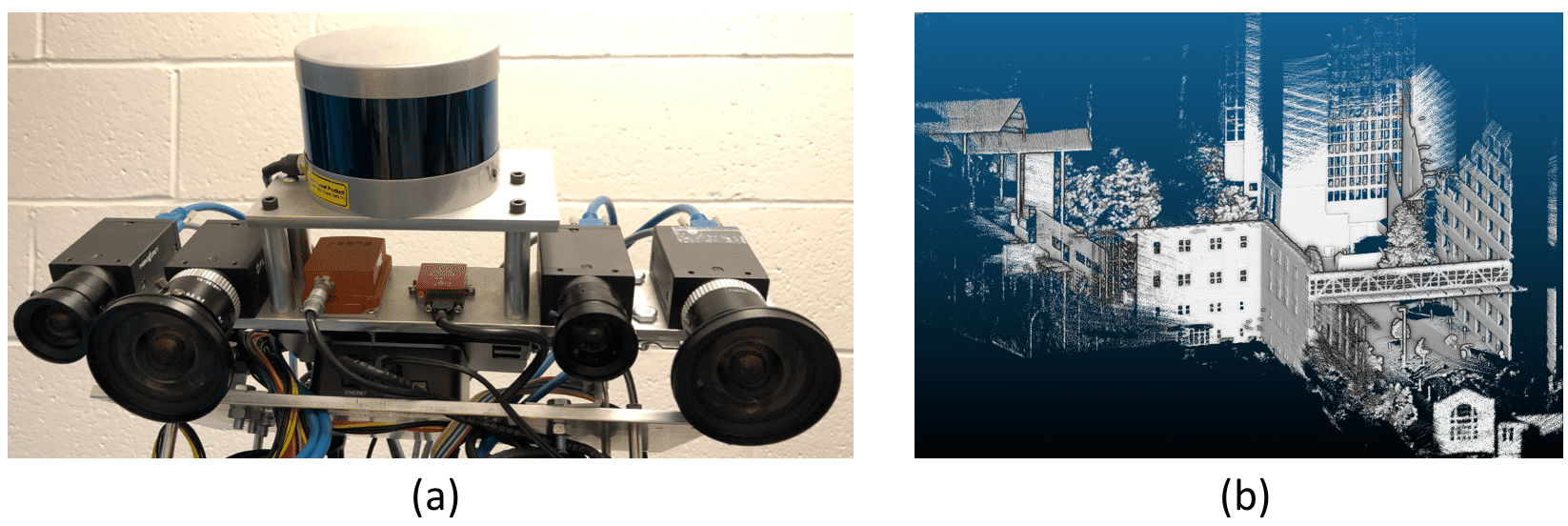}
    \caption{(a) Experimental platform built. (b) Mapping result from an outdoor test. Streetlight is reconstructed clearly.}
    \label{platformFig}
    \vspace{-2mm}
\end{figure}



\begin{figure*}[t!]
    \centering
    \includegraphics[scale=0.5]{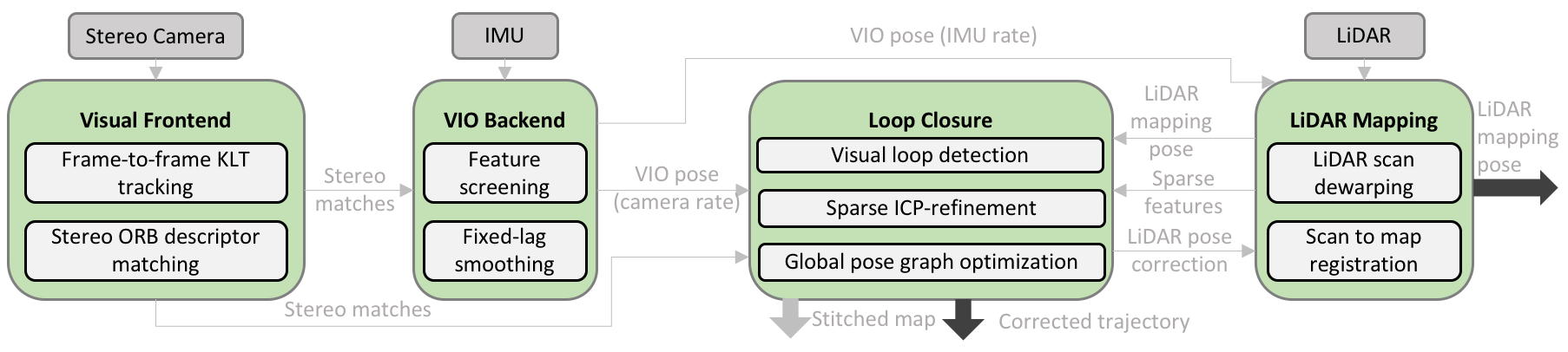}
    \caption{The system diagram of VIL-SLAM. Sensors are in gray and modules are in green. Arrows indicate how messages flow within the system. The dark thick arrows indicate the system real-time output and the light thick arrow indicates the output generated in post-processing near real-time.}
    \label{system_overview_block_diagram}
    \vspace{-4mm}
\end{figure*}

\section{Related work}\label{secRW}
Current VIO literature introduces various formulations to integrate visual and inertial data. The literature characterizes different approaches into \textit{tightly-coupled system} \cite{refVINSMONO, refOKVIS, Hsiung18iros}, in which visual information and inertial measurements are jointly optimized, or \textit{loosely-coupled system} \cite{refDSO, refPTAM, ARMMFA, RTOVI}, in which IMU is a separate module and fused with a vision-only state estimator. The approaches could be further divided into either filtering-based \cite{refRSVIOupenn, ref11, refROVIO, Wu2015ASR, RTOVI, 6385458} or graph-optimization based \cite{refVINSMONO, refOKVIS, Hsiung18iros, Indelman2013InformationFI, DVISO}. Tightly-coupled optimization-based approaches, taking the benefit of minimizing residuals iteratively, usually achieve better accuracy and robustness with a higher computation cost. In our work, we bound the computation cost by forming landmarks in a structureless fashion and only optimizing for a fixed-size pose graph to achieve the real-time performance. 

Current state-of-the-art SLAM systems using just laser scanner are \cite{7353456, 7487648, 7279468, refLOAM, IMLSSLAM}, in which a motion model is required, either a constant velocity model or a Gaussian process. Approach in \cite{6907626} combines stereo cameras and a laser scanner. It has motion estimation generated from a visual odometry (VO) and refined by matching laser scans. The differences to our system are that they use multi-resolution grid map representation and ours uses sparse point cloud to localize and outputs dense point cloud. Also, VIO is usually more robust and accurate compared to a VO \cite{Delmerico2018ABC}. VLOAM \cite{refVLOAM}, which uses an IMU, a monocular camera, and a laser scanner is the most similar existing system to ours. One difference is that we use a tightly-coupled VIO as the motion model to initialize the LiDAR mapping algorithm whereas VLOAM uses loosely-coupled IMU and camera. Though our VIO is more robust, VLOAM has a more interactive system where information from both camera and LiDAR module could be used for IMU biases correction. One addition that VIL-SLAM has is the LiDAR enhanced loop closure.

\section{System overview}\label{secSO}
The system has four modules as shown in Fig. \ref{system_overview_block_diagram}. The visual frontend takes stereo pairs from the stereo cameras. It performs frame to frame tracking and stereo matching, and outputs stereo matches as visual measurements. The stereo VIO takes stereo matches and IMU measurements, performs IMU pre-integration and tightly-coupled fixed-lag smoothing over a pose graph. This module outputs VIO pose at IMU rate and camera rate. LiDAR mapping module uses the motion estimate from the VIO and performs LiDAR points dewarping and scan to map registration. The loop closure module conducts visual loop detection and initial loop constraint estimation, which is further refined by a sparse point cloud ICP alignment. A global pose graph constraining all LiDAR poses is optimized incrementally to obtain a globally corrected trajectory and a LiDAR pose correction in real-time. They are sent back to LiDAR mapping module for map update and re-localization. In post processing, we stitch the dewarped LiDAR scans with the best estimated LiDAR poses to have the dense mapping results (Fig. \ref{traj_map_res}).

\section{Visual frontend}\label{secVF}
Visual frontend accepts a stereo pair, and performs frame to frame tracking and stereo matching for the generation of a set of stereo-matched sparse feature points, namely, stereo matches. A stereo match could either be one tracked from previous stereo pair, or a new one extracted in this pair. The frame to frame tracking performance directly affects the temporal constraints quality while the stereo matching helps constrain the scale. These two tasks are crucial for any stereo visual odometry. Direct methods show robust and efficient temporal tracking results in recent years \cite{refDSO, refSVO}. Thus, we use Kanade Lucas Tomasi (KLT) feature tracker \cite{refLKT} to track all feature points in the previous stereo matches, either in the left or right image. Only when they are both tracked, we have a tracked stereo match and it is pushed into the output. Large stereo baseline helps scale estimation and reduces degeneracy issues caused by distant features. We use feature-based methods which are better suited to handle large baselines than KLT. If the number of tracked stereo matches is below a threshold, we perform feature extraction using Shi-Tomashi Corner detector \cite{refSHI}, followed by a feature elimination process in which features that have pixel coordinate distance to any existing features smaller than a threshold are deleted. ORB (Oriented FAST and Rotated BRIEF) \cite{refORB} descriptors are then computed on all survived features, followed by a brute-force stereo matching to obtain new stereo matches. The system initializes by performing stereo matching on the first stereo pair.

\section{Stereo visual inertial odometry}\label{secSVIO}
The goal of the stereo VIO is to provide real-time accurate state estimate at a relatively high frequency, serving as the motion model for the LiDAR mapping algorithm. A tightly-coupled fixed-lag smoother operating over a pose graph is a good trade-off between accuracy and efficiency. Optimization-based methods in general allow for multiple re-linearization to approach the global minimum. A fixed-lag pose graph optimizer further bounds the maximum number of variables, and hence the computation cost is bounded. Since bad visual measurements cause convergence issues, we enforce a strict outlier rejection mechanism on visual measurements. The system eliminates outliers by checking the average reprojection error, both stereo and temporal.

\begin{figure}[t!]
    \centering
    \includegraphics[scale=0.18]{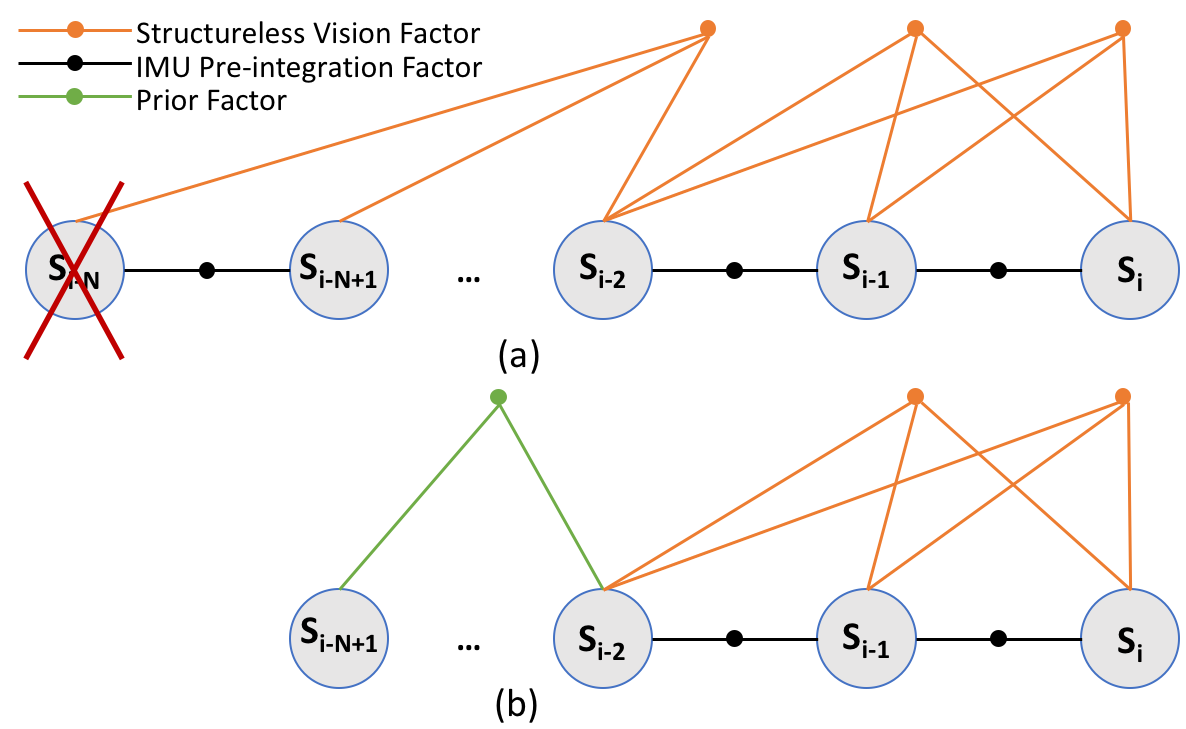}
    \caption{Fixed-lag pose graph formulation in the VIO. State variables being optimized are circled, where $i$ stands for the current state and $N$ is the window size. (a) The state to be marginalized is crossed. (b) After marginalization, prior factors are added back on related variables.}
    \label{VIO_pose_graph}
    \vspace{-2mm}
\end{figure}

The VIO proposed has \textit{IMU Pre-integration Factor} and \textit{Structureless Vision Factor} as constraints. The graph representation is shown in Fig. \ref{VIO_pose_graph}. Variables to be optimized are the states inside the window. Denote $\textbf{S}_t$ as the state variable at the stereo frame time $t$. $\textbf{S}_t$ contains the 6 Degrees of Freedom (DoF) system pose $\xi_t$ (IMU frame), the associated linear velocity $\textbf{v}_t$, accelerometer bias $\textbf{b}^a_t$, and gyroscope bias $\textbf{b}^g_t$. The window of state variables being estimated are of the most recent $N$ stereo frames. Past state variables are marginalized, producing prior factors on related variables.

\subsection{IMU pre-integration factor}

We follow the IMU pre-integration method \cite{refIMUPre}\cite{refIMUPre0} to generate relative IMU measurements between $\textbf{S}_i$ and $\textbf{S}_j$. Using the pre-integration technique, re-linearization could be performed efficiently during optimization. The residual represented by the IMU pre-integration factor is $\textbf{r}^I_{ij}$, which consists of three terms: the residual of pose ($\textbf{r}_{\Delta\xi ij}$), velocity ($\textbf{r}_{\Delta\textbf{v}ij}$), and biases ($\textbf{r}_{\Delta\textbf{b}ij}$).

\subsection{Structureless vision factor}
Visual measurements are modeled in a structureless fashion, similar to \cite{refIMUPre}\cite{refIMUPre2}\cite{refIMUPre3}. Consider a landmark $p$, whose position in global frame is $\textbf{x}_p\in\mathbb{R}^3$, is observed by multiple states and denote the set of states observing $p$ as $\{\textbf{S}\}_p$. For any state $\textbf{S}_k$ in $\{\textbf{S}\}_p$, denote the residual formed by measuring $p$ as in the left camera image as $\textbf{r}^V_{\xi_{k,lc},p}$ ($\xi_{k,lc}$ is the left camera pose, obtained by applying a IMU-camera transformation to $\xi_k$):
\begin{equation} \label{visualResidual}
    \textbf{r}^V_{\xi_{k,lc},p} = \textbf{z}_{\xi_{k,lc},p}-h(\xi_{k,lc},\textbf{x}_p)
\end{equation}
where $\textbf{z}_{\xi_{k,lc},p}$ is the pixel measurement of $p$ in the image and $h(\xi_{k,lc},\textbf{x}_p)$ encodes a perspective projection. Same formulation is derived for the right camera image. Iterative methods are adopted for optimizing the pose graph, and hence linearization of the above residual is required. Equation (\ref{linearizedVR}) shows the linearized residuals for landmark $p$.
\begin{equation}\label{linearizedVR}
    \sum_{S_p}||\textbf{F}_{kp}\delta \xi_k + \textbf{E}_{kp}\delta\textbf{x}_p + \textbf{b}_{kp}||^2
\end{equation}
where the Jacobians $\textbf{F}_{kp}$, $\textbf{E}_{kp}$ and the residual error $\textbf{b}_{kp}$ are results from the linearization and normalized by $\Sigma^{1/2}_c$, the visual measurement covariance. Stacking each individual component inside the sum into a matrix we have 
\begin{equation}
    ||\textbf{r}^V_p||^2_{\Sigma_C} = ||\textbf{F}_{p}\delta \xi_k + \textbf{E}_{p}\delta x_p + \textbf{b}_{p}||^2
\end{equation}
To avoid optimizing over $\textbf{x}_p$, we project the residual into the null space of $\textbf{E}_{p}$: Premultiply each term by $\textbf{Q}_p\doteq\textbf{I}-\textbf{E}_{p}(\textbf{E}_{p}^\top\textbf{E}_{p})^{-1}\textbf{E}_{p}^\top$, an orthogonal projector of $\textbf{E}_{p}$ \cite{refIMUPre}. We thus have the \textit{Structureless Vision Factor}, for landmark $p$ as
\begin{equation}
    ||\textbf{r}^V_p||^2_{\Sigma_C} = ||\textbf{Q}_p\textbf{F}_{p}\delta \xi_k + \textbf{Q}_p \textbf{b}_{p}||^2
\end{equation}
\subsection{Optimization and marginalization}
Given the residuals, the pose graph optimization is a \textit{maximum a posteriori} (MAP) problem whose optimal solution is
\begin{equation}
    \textbf{S}^*_w = \arg\min_{S^*_w}(||\textbf{r}_0||^2_{\Sigma_0} +\sum_{i\in w} ||\textbf{r}^I_{i(i+1)}||^2_{\Sigma_{I}} + \sum_{p}||\textbf{r}^V_p||^2_{\Sigma_{C}})
\end{equation}
where $\textbf{S}^*_w$ is the set of state variables inside the window. $\textbf{r}_0$ and $\Sigma_0$ are prior factors and their associated covariance. $\Sigma_I$ is the covariance of the IMU measurements. We use the Levenberg-Marquart optimizer to solve this nonlinear optimization problem. The most recent $N$ state variables are maintained inside the optimizer. Schur-Complement marginalization \cite{schurComplement} is performed on state variables getting out of the window. Prior factors are then added to related variables inside the window as in Fig. \ref{VIO_pose_graph}(b).

\begin{figure}[t!]
    \centering
    \includegraphics[scale=0.18]{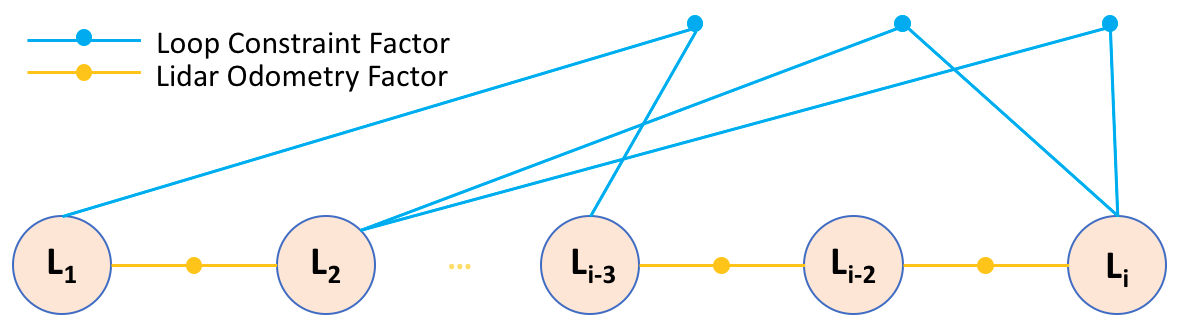}
    \caption{The global pose graph consists of the \textit{LiDAR Odometry Factor} and the \textit{Loop Constraint Factor}. $i$ stands for the current scan.}
    \label{global_pose_graph}
    \vspace{-2mm}
\end{figure}

\section{Lidar mapping}\label{secLSTMM}
LiDAR mapping uses high frequency IMU rate VIO poses as the motion prior to perform LiDAR points dewarping and scan to map registration. Denote a scan $\chi$ as the point cloud obtained from one complete LiDAR rotation. Geometric features including points on sharp edges and planar surfaces are extracted from $\chi$ before dewarping \cite{refLOAM, refVLOAM}. The registration is then based on feature points from current scan to the map (all previous feature points), solved as an optimization problem by minimizing Euclidean distance residuals formed by the feature points as in \cite{refLOAM}.

\subsection{LiDAR scan dewarping}

Dewarping is required as points from a LiDAR scan are timestamped differently. Denote any time within a scan as $t_{i}$. We dewarp all points to the time of end of scan $t_{k+1}$ based on IMU rate VIO poses. Denote a LiDAR point at $t_i$ as $\textbf{P}_{i}$ and the dewarped itself as $\tilde{\textbf{P}}_i$, we have  
\begin{equation}
    \tilde{\textbf{P}}_{i}=(\textbf{T}_{k+1}^L)^{-1}\textbf{T}_{i}^{L}\textbf{P}_{i}
\end{equation}
where $\textbf{T}_{k+1}^{L}$, $\textbf{T}_{i}^{L}$ are LiDAR frame poses transformed from the closest IMU rate VIO poses.

\subsection{Scan to map registration}
Feature points from the dewarped scan $\tilde{\chi}$ are registered to the map, optimizing for the LiDAR mapping pose at $t_{k+1}$ denoted as $\textbf{L}_{k+1}$. Denote the initial estimate of $\textbf{L}_{k+1}$ as $\textbf{L}_{k+1}^*$, we have:
\begin{equation}
    \textbf{L}_{k+1}^* = \textbf{L}_k\textbf{T}_{trans}^{L}   
\end{equation}
where $\textbf{L}_k$ is the optimized previous LiDAR mapping pose and $\textbf{T}_{trans}^{L}$ is the relative transformation obtained based on IMU rate VIO poses. All dewarped feature points are then transformed to world coordinate system by $\textbf{L}_{k+1}^*$ for registration.

The residual $\textbf{r}_E$ of an edge feature point in the current scan, is the Euclidean distance between itself and the line formed by the two closest edge points in the map. The residual $\textbf{r}_{U}$ of a surface point in the current scan is the distance between itself and the planar patch formed by the three closest surface points in the map. \cite{refLOAM} Incorporating $\textbf{L}_{k+1}^*$, we can rewrite the two residuals as:
\begin{equation}
f_{E}( E_{(c,i)}^{L}, \textbf{L}_{k+1}^*) = \textbf{r}_{E}
\end{equation}
\begin{equation}
f_{U}( U_{(c,i)}^{L}, \textbf{L}_{k+1}^*) = \textbf{r}_{U}
\end{equation}
where $E_{(c,i)}^{L}$ and $U_{(c,i)}^{L}$ are the 3D position of the $i$th dewarped feature point in the LiDAR coordinate system. Levenberg-Marquardt optimizer is used to solve this nonlinear optimization problem, formed by stacking the cost functions for all feature points.

\begin{figure}[!t]
\captionsetup[subfloat]{farskip=1pt,captionskip=1pt}
\subfloat[Highbay]{\includegraphics[width=\columnwidth]{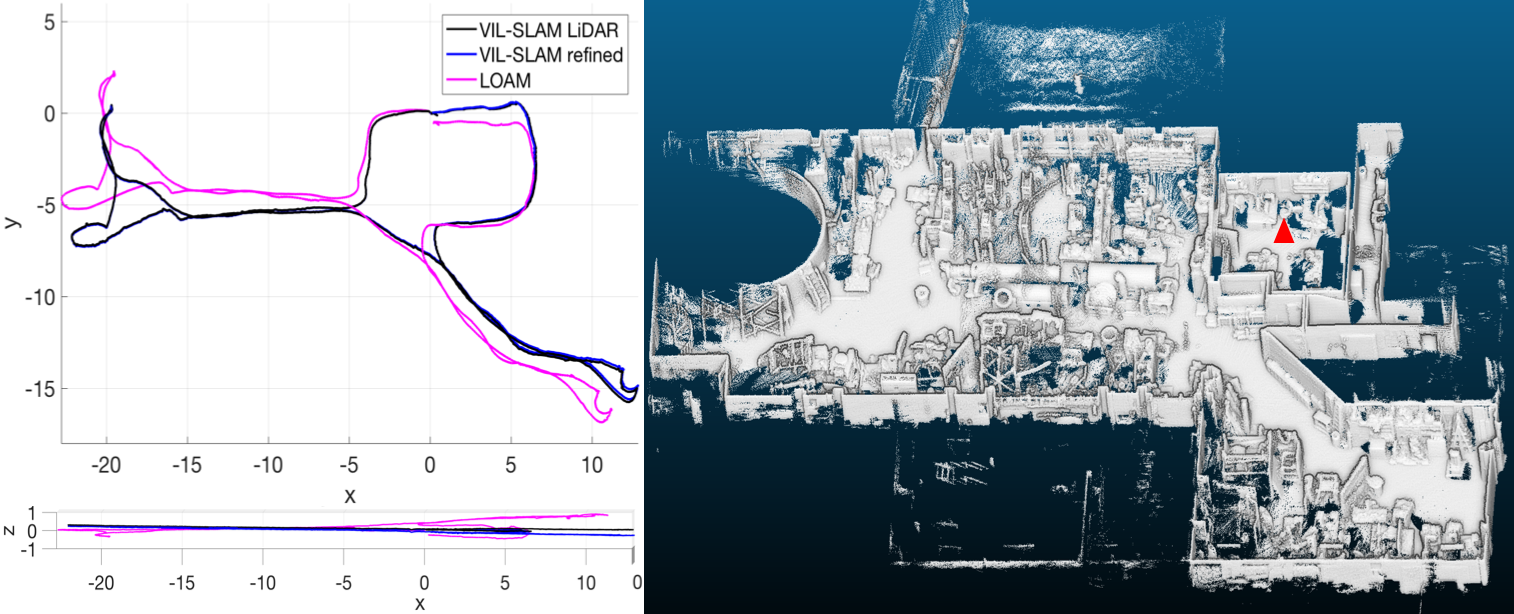}}

\subfloat[Hallway]{\includegraphics[width=\columnwidth]{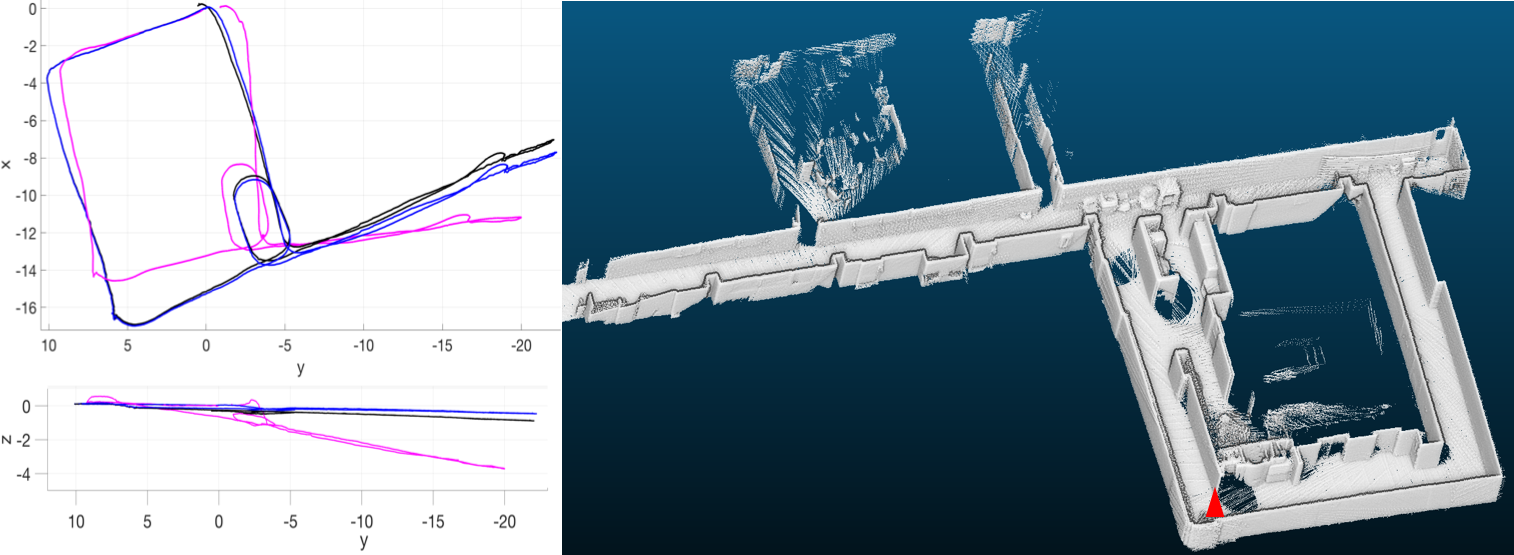}}

\subfloat[Tunnel]{\includegraphics[width=\columnwidth]{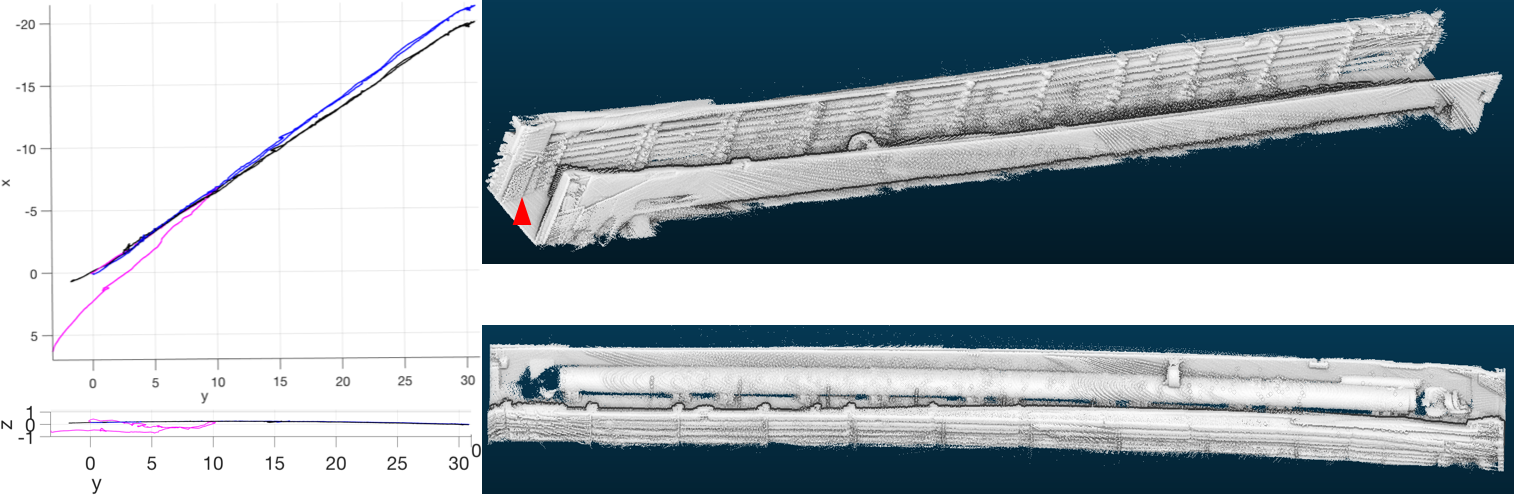}}

\subfloat[Huge Loop]{\includegraphics[width=\columnwidth]{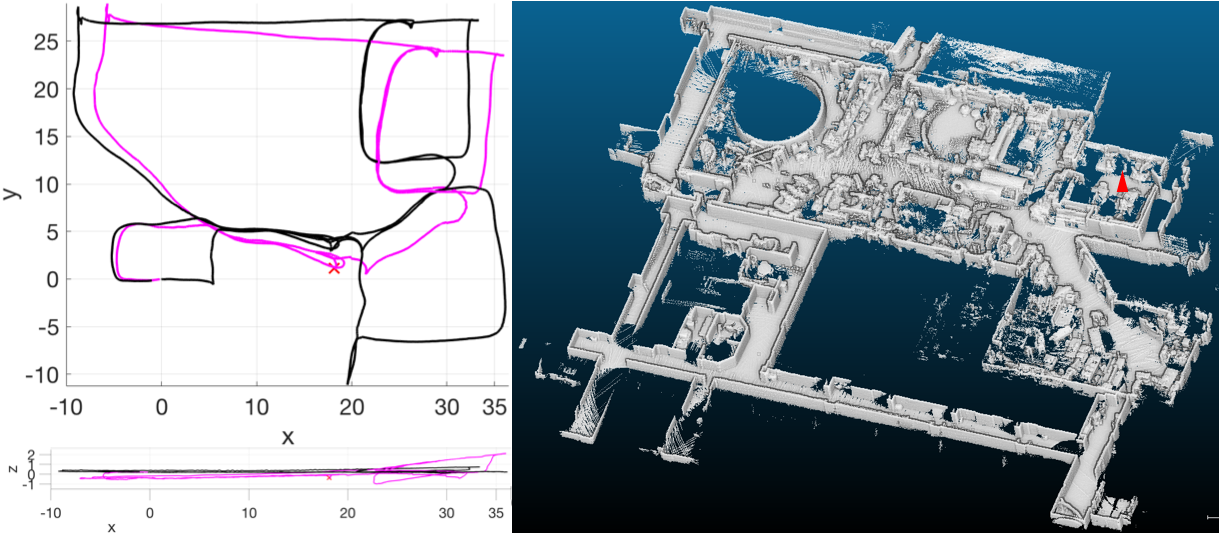}}

\subfloat[Outdoor]{\includegraphics[width=\columnwidth]{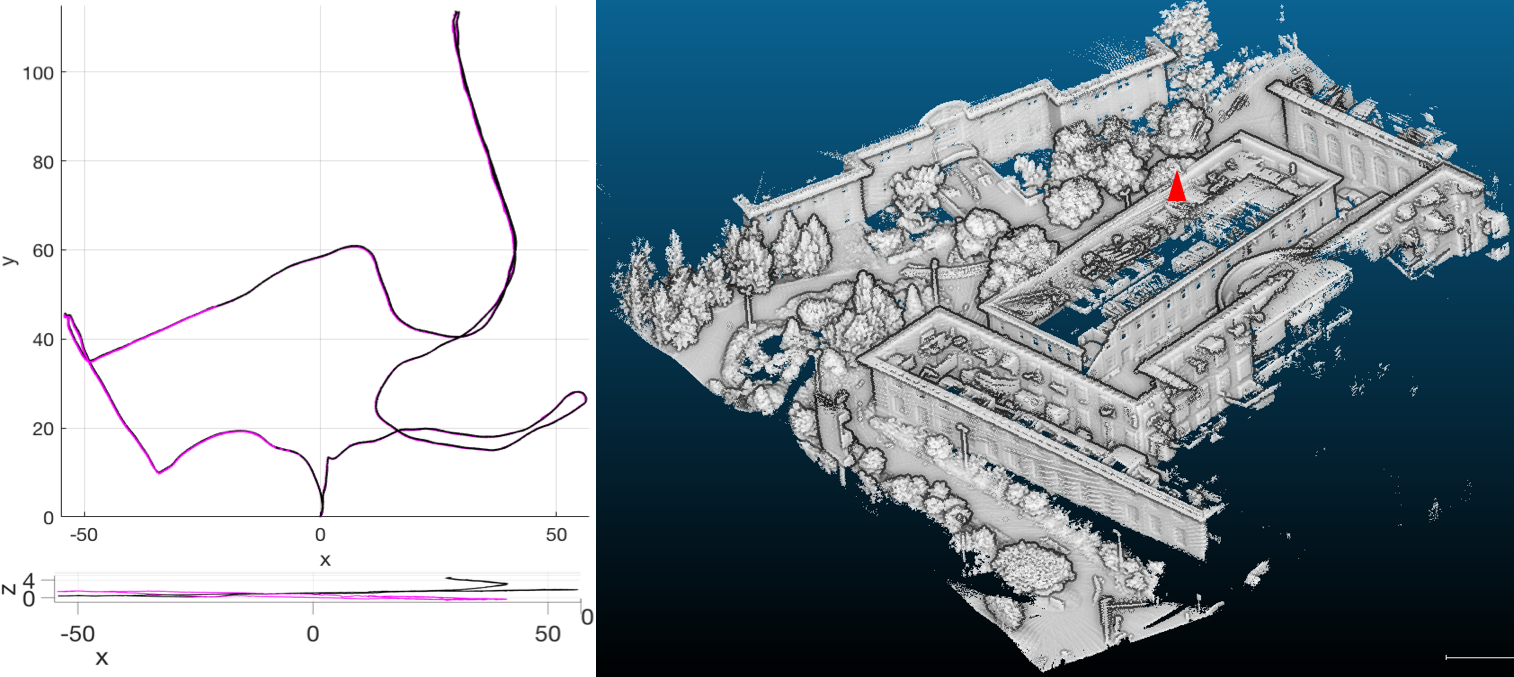}}

\caption{Trajectories from VIL-SLAM and LOAM are shown on the left and maps generated by VIL-SLAM are shown on the right. Start(end) position is labeled with red triangle in the map and is the origin in the plot.}
\label{traj_map_res}
\vspace{-4mm}
\end{figure}

\section{Lidar enhanced loop closure}\label{secSIELC}
Loop closure is critical to any SLAM system as long term operation introduces drift. The objective of loop closure is to eliminate drift by performing a global pose graph optimization which incorporates loop constraints and relative transformation information from LiDAR mapping. To better assist LiDAR mapping, the corrected LiDAR pose is sent back in real-time so that feature points from new scans are registered to the revisited map. We propose adding ICP alignment in addition to visual Bag-of-Words \cite{refBoW} loop detection and PnP loop constraint formulation. The system uses iSAM2 \cite{refiSAM2}, an incremental solver, to optimize the global pose graph, achieving real-time performance. 

\subsection{Loop detection}
Stereo images and LiDAR scans are associated using their timestamps. Let us denote these as key images and key scans respectively. To prevent false loop detection we restrict candidates within a certain time threshold. Loop candidates are detected by testing the key images with the Bag-of-Words \cite{refBoW} database of previous key images. Furthermore, We match feature descriptors of the left key image with the loop candidates to filter out the false positives.

\subsection{Loop constraint}

The system first obtains visual loop constraint as an initial estimate. Since we use a structureless formulation for visual landmarks, triangulation on all the stereo matched features in the loop candidate is performed to obtain their 3D location. Their associations to current key images are given by descriptor match. The visual loop constraint is then evaluated using EPNP \cite{refEPNP}. To improve the accuracy of the visual loop constraint, we use ICP alignment on the feature points of the corresponding LiDAR key scans. With a bad initialization or a larger point count, ICP takes longer to converge and consumes more computation resources. However, the visual loop constraint provides a good initialization point and the ICP only uses sparse feature points (Section \ref{secLSTMM}), which makes it converge faster.

\subsection{Global pose graph optimization}

The graph representation of the global pose graph is shown in Fig. \ref{global_pose_graph}. It contains all the available LiDAR mapping poses as variables, constrained by the \textit{LiDAR Odometry Factor} and the \textit{Loop Constraint Factor}, both are measurements of the relative transformation: $(\textbf{L}_u)^{-1}\textbf{L}_v$ where $u$ and $v$ stand for scan ID and $\textbf{L}_u$, $\textbf{L}_v$ are the associated poses. For the \textit{LiDAR Odometry Factor}, $u$ is the previous scan ID. For the \textit{Loop Constraint Factor}, $u$ is the key scan ID found as loop. For both cases, $v$ is the current scan ID. Poses are expressed in 6 DoF minimum form in the optimization. To realize real-time performance, we use iSAM2 \cite{refiSAM2} to incrementally optimize the global pose graph.

\subsection{Re-localization}
Once a true loop closure candidate is found, LiDAR mapping buffers the feature points (without registering them to the map) until it receives loop correction. The loop correction contains globally optimized trajectory. LiDAR mapping updates its map, adds the buffered feature points to the map and then resumes its operation. We can afford to update the map in real-time because (a) loop closure has a real-time performance (b) the sparse feature map does not take much memory, and (c) scan to map registration is fast enough to catch up the LiDAR data rate.

\begin{table}[!t]
\captionsetup{skip=0pt}
\caption{FDE (\%) and MRE (m) TEST RESULTS\label{FDE}}
\begin{center}
\begin{tabular}{|c|c||c|c||c|c|}
\hline
 \multirow{2}{*}{Test} & Total & \multicolumn{2}{|c||}{FDE} & \multicolumn{2}{|c|}{MRE} \\
 \cline{3-6}
 & Length & VIL-SLAM & LOAM & VIL-SLAM & LOAM \\
\hline
\textit{Highbay} & 118 & \textbf{0.08} & 0.56 & \textbf{0.08} & 0.22 \\
\hline
\textit{Hallway} & 103 & \textbf{0.61} & 0.91 & \textbf{0.10} & 0.27 \\
\hline
\textit{Tunnel} & 85 & \textbf{1.86} & -\tablefootnote{"-" indicates not finished. "$\times$" indicates missing data.} & $\times$ & $\times$ \\
\hline
\textit{Huge} & \multirow{2}{*}{318} & \multirow{2}{*}{\textbf{0.01}} & \multirow{2}{*}{-} &\multirow{2}{*}{\textbf{0.22}} & \multirow{2}{*}{0.36} \\
\textit{Loop} & & & & & \\
\hline
\textit{Outdoor} & 528 & 0.02 & 0.02 & $\times$ & $\times$ \\
\hline
\end{tabular}
\end{center}
\vspace{-4mm}
\end{table}

\section{Experimental results}\label{secER}
We evaluate VIL-SLAM and compare it with the best real-time LiDAR based system, LOAM\footnote{This is the best implementation of LOAM we could find online https://github.com/laboshinl/loam\_velodyne}\cite{refLOAM} on custom datasets. We did not use KITTI odometry dataset \cite{Geiger2012CVPR} because their evaluation sequences do not have inertial measurements which are needed for VIO. Also, most KITTI sequences are not challenging. So they do not evaluate the robustness of these systems which is the main focus of our experiments. We also evaluate the stereo VIO submodule (VIL-VIO) using the EuRoC MAV dataset \cite{refEUROC}.

\begin{figure}[!t]
\captionsetup[subfloat]{farskip=1pt,captionskip=1pt}
\subfloat[Highbay]{\includegraphics[width=\columnwidth]{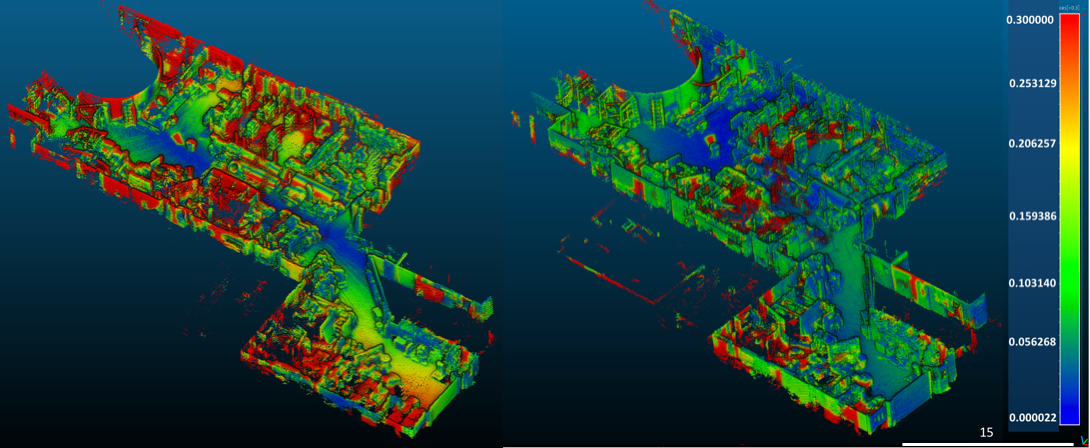}}

\subfloat[Hallway]{\includegraphics[width=\columnwidth]{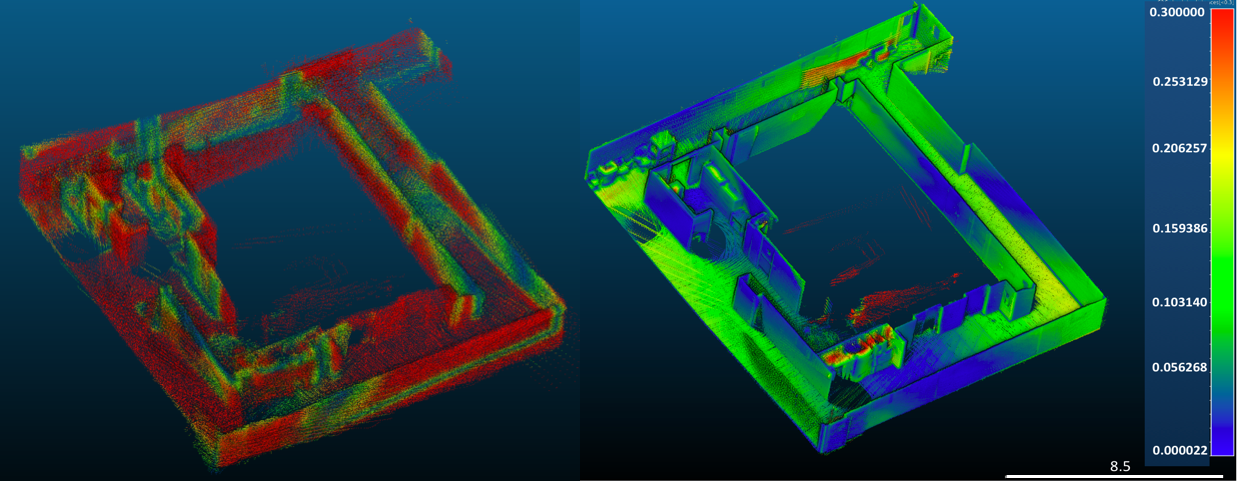}}

\subfloat[Huge Loop]{\includegraphics[width=\columnwidth]{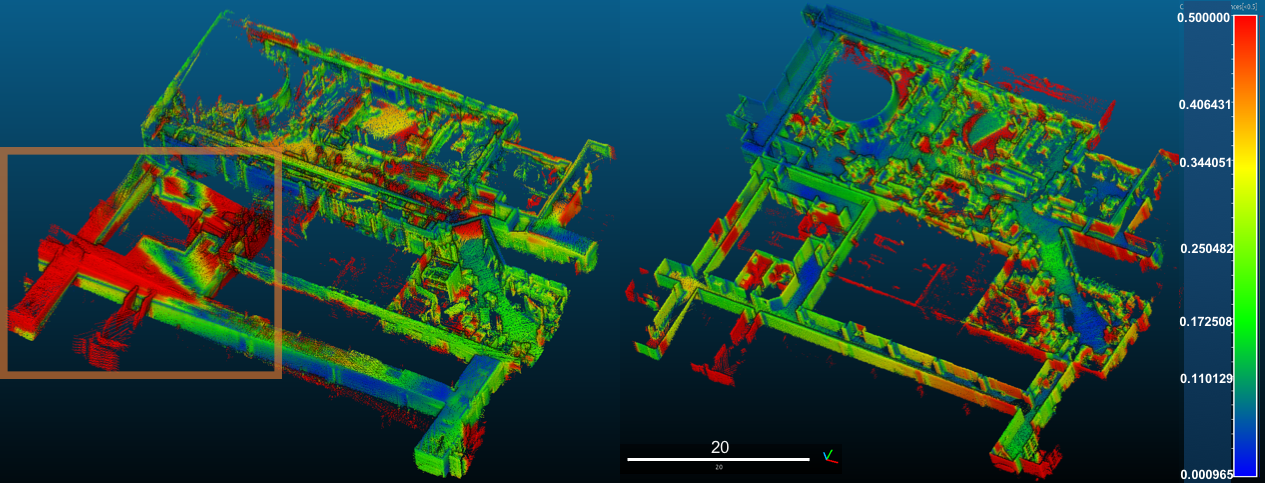}}

\caption{Map registration error of VIL-SLAM (right) and LOAM (left) comparing to the model. Errors above 0.3m are colored red for (a-b) and 0.5m for (c). Discontinuous red regions inside the blue and green are due to lack of the model caused by occlusions of the Faro scans.}
\label{map_comp_res}
\vspace{-2mm}
\end{figure}

\subsection{Platform and software}
We built a platform (Fig. \ref{platformFig}(a)) with two megapixel cameras, a 16 scan-line LiDAR, an IMU (400Hz), and a 4GHz computer (with 4 physical cores). We built a custom microcontroller based time synchronization circuit that synchronizes the cameras, LiDAR, IMU and computer by simulating GPS time signals. The software pipeline is implemented in C++ with ROS communication interface. We use GTSAM library \cite{refGTSAM} to build the fixed-lag smoother in the VIO. For loop closure, we use ICP module from  LibPointMatcher \cite{refLibPointMatcher} to align point clouds, DBoW3 \cite{refDBoW3} to build the visual dictionary, and iSAM2 \cite{refiSAM2} implementation in GTSAM \cite{refGTSAM} to conduct global optimization. 

\begin{figure}[t!]
    \captionsetup{captionskip=1pt}
    \includegraphics[width=\columnwidth]{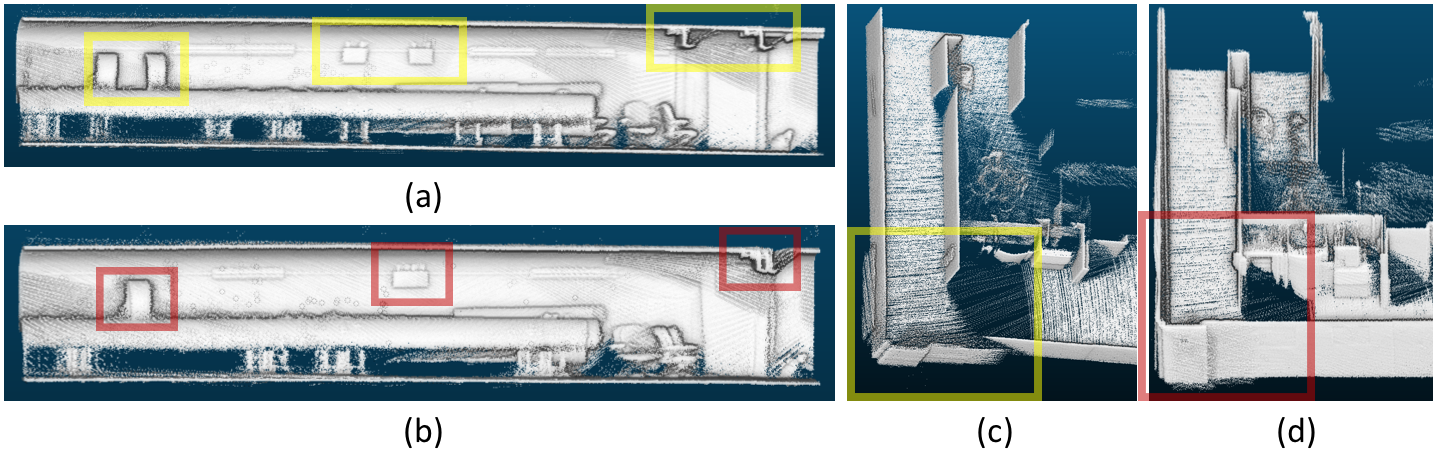}
    \caption{(a) Map of the tunnel stitched using LIDAR mapping poses. (b) Map of the tunnel stitched using globally refined poses. Double image in (a) is mostly eliminated but not fully, because only one loop constraint is generated, not enough for a full correction. (c) Map of the hallway stitched using LIDAR mapping poses. (d) Map of the hallway stitched using globally refined poses. Double image in (c) is mostly eliminated. Walls are aligned with two loop constraints.}
    \label{lccc}
    \vspace{-2mm}
\end{figure}

\subsection{Tests and results}
We present results from five representative environments including featureless hallways, cluttered highbays, tunnels, and outdoor environments. The data collection started and ended at the same point for all these sequences. Odometry (LiDAR mapping pose) is evaluated based on the final drift error (FDE). Mapping results are evaluated in terms of mean registration error (MRE) using Faro scans as ground truth. We first align the map with the model (Faro scans), and then compute the Euclidean distance between a map point and its closest point in the model \cite{refCC}. The odometry FDE and mapping results are shown in Table \ref{FDE} with the better ones in bold. The trajectories and cross-sectioned maps are shown in Fig. \ref{traj_map_res}. The map comparisons are shown in Fig. \ref{map_comp_res}.

The \textit{highbay} is an indoor warehouse which is open, structured, and rich in features. However, frequent structural occlusions could be a challenge for the visual frontend and the LiDAR feature extraction part. Both VIL-SLAM and LOAM handle this environment pretty well. For VIL-SLAM, LiDAR mapping module registers most of its scan to map, largely reducing the odometry error. Loop closure recognizes the starting position and closes the loop. The map is generated using the globally refined poses, with the majority of map errors below 0.15m.

The \textit{hallway} and \textit{tunnel} tests are challenging environments because of lack of visual features and the degeneracy issue along traversal direction for LiDAR. LOAM accumulates large error in the hallway, and fails the tunnel test mainly due to the degeneracy issue. Aided by the stereo VIO module (VIL-VIO), VIL-SLAM succeeds both tests. In the \textit{hallway} test, the visual frontend returns fewer reliable measurements because of the featureless walls, under-constraining the VIO. This corrupts the map as observed by wall misalignment, which is later corrected by loop closure as shown in Fig. \ref{lccc}(c-d). Loop closure detects the loop twice when approaching the endpoint, lowering FDE to 0.05\% and generating a refined map. In the \textit{tunnel} test, because of the degeneracy issue, VIL-SLAM struggles as well and accumulates some error in the traversal direction. However, loop closure detects the loop at about 3m from the end point, lowering the FDE down to 0.08\% and correcting the map as shown in Fig. \ref{lccc}(a-b).

The \textit{huge loop} test features challenges from both \textit{hallway} and \textit{highbay} environments. In addition, we end the trajectory by re-entering the highbay after traversing along a long narrow corridor. LOAM fails this test after re-entering the highbay, at the place labeled by a red cross in Fig. \ref{traj_map_res}(d). We think this is because it fails to register new scans to the original \textit{highbay} map caused by a large error in z-direction accumulated in the corridor. VIL-SLAM succeeds in this test. Without loop closure being triggered, it achieves 0.01\% FDE in odometry. VIL-SLAM is robust and achieves this result by successfully registering new scans to the original \textit{highbay} map at re-entry. The map generated with the odometry estimate of VIL-SLAM is compared with the map generated with LOAM before its failure. The boxed region is where LOAM accumulates errors leading to its failure.

The \textit{outdoor} test features an outdoor trajectory which is 546m long and includes a gentle slope. Pedestrians and cars were observed which served as potential outliers. VIL-SLAM and LOAM have comparable results along the xy-plane. However, LOAM fails to capture the changes in the z-direction. The inaccuracy in z of LOAM is also observed in the previous tests.

Overall, VIL-SLAM generates more accurate mapping results and lower FDE compare to LOAM when they both finish. Also, VIL-SLAM succeeds the more challenging environments where LOAM fails with qualitatively good mapping and odometry results.

\begin{figure}[t!]
    \centering
    \includegraphics[scale=0.15]{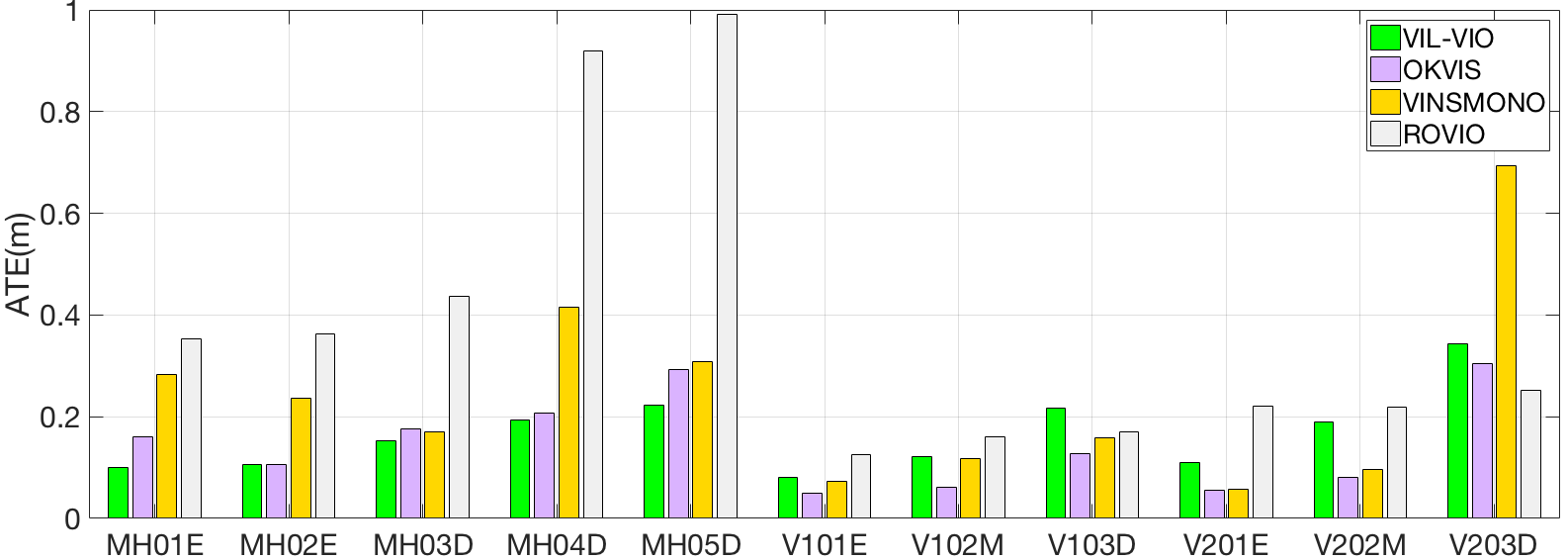}
    \caption{Root mean square error of ATE for EoRoC Dataset.}
    \vspace{-3mm}
    \label{eurocFig}
\end{figure}

\subsection{EuRoC MAV Dataset test}
VIL-VIO contributes to the robustness and accuracy of VIL-SLAM. We evaluate the VIO using the EuRoC MAV dataset \cite{refEUROC} in terms of the absolute trajectory error (ATE) as in \cite{refATE}. Fig. \ref{eurocFig}\footnote{A sequence is named in the first four letters and the difficulty level is encoded in the last letter (E:easy, M:medium, D:difficult)} shows the comparison results between VIL-VIO and three state-of-the-art methods. Results for VIL-VIO are deterministic, obtained in real-time on a desktop with 3.60GHz i7-4790 CPU. Results for the other methods are the better ones from experiments in \cite{Hsiung18iros} and \cite{refRSVIOupenn}. VIL-VIO succeeds all sequences with accuracy comparable with the others, verifying its capability to handle aggressive motion, illumination changes, motion blur and textureless regions.

\section{Conclusions}

VIL-SLAM is a state-of-the-art odometry and mapping system designed to robustly operate long term in different environments. Current framework loosely couples VIL-VIO and LiDAR mapping. We are extending it to a tightly-coupled framework such that refined pose estimate from LiDAR mapping could be used for IMU biases correction. In loop closure, ICP refinement operates on sparse feature points between scans. We suspect that we would obtain a better loop constraint by matching a scan to map.

\addtolength{\textheight}{-12cm}   

\bibliography{main.bib}
\bibliographystyle{ieeetr}
\end{document}